\documentclass[11pt]{article} 
\usepackage[english]{babel}
\usepackage[utf8]{inputenc}
\usepackage[T1]{fontenc}
\usepackage{latexsym}
\usepackage{url}
\usepackage[dvips]{graphicx}
\usepackage[normalem]{ulem}
\usepackage{amsmath,amsfonts,amssymb,amsthm,bm}
\usepackage{psfrag}
\usepackage{fullpage}
\usepackage{subcaption}
\usepackage{enumitem}
\usepackage{color}
\usepackage[table]{xcolor}
\usepackage[font={small}]{caption}
\usepackage[output-decimal-marker={,},exponent-product=\cdot]{siunitx}
\usepackage[hang,flushmargin]{footmisc}
\usepackage{hyperref}

\usepackage{booktabs}
\usepackage[section]{placeins}
 \usepackage[ruled, vlined]{algorithm2e}
\usepackage{algorithmic}
\usepackage{microtype}
\usepackage{xspace}
\usepackage{complexity}
\usepackage{mathtools, nccmath}
\usepackage[misc]{ifsym}
 \usepackage{multirow}

\newcommand{\noi}{\noindent}

\mathchardef\mhyphen="2D

\def\noi{\noindent}

\numberwithin{equation}{section}

\usepackage{mleftright} 

\small \normalsize

\begin{document}

\title{Solving the Order Batching and Sequencing Problem using Deep Reinforcement Learning}

\author{
{Bram Cals} \footnotemark[1]
\and {Yingqian Zhang (\Letter)}\thanks{
 Eindhoven University of Technology, Eindhoven, The Netherlands, e-mail: {yqzhang@tue.nl}} 
\and {Remco Dijkman} \footnotemark[1]
\and {Claudy van Dorst} \footnotemark[2]\thanks{Vanderlande Industries B.V., Veghel, The Netherlands}
}

\date{This version: 16 June 2020}

\maketitle
\sloppy

\begin{abstract} 

\noi  In e-commerce markets, on time delivery is of great importance
to customer satisfaction. 
In this paper, we present a 
Deep Reinforcement Learning (DRL) approach for deciding how and when orders should be batched and picked in a warehouse to minimize the number of tardy orders. In particular, the technique facilitates making decisions on whether an order should be picked individually (pick-by-order) or picked in a batch with other orders (pick-by-batch), and if so with which other orders. We approach the problem by formulating it as a semi-Markov decision process and develop a vector-based state representation that includes the characteristics of the warehouse system. This allows us to create a deep reinforcement learning solution that learns a strategy by interacting with the environment and solve the problem with a proximal policy optimization algorithm. 
We evaluate the performance of the proposed DRL approach by comparing it with several batching and sequencing heuristics in different problem settings. The results show that the DRL approach is able to develop a strategy that produces consistent, good solutions and performs better than the proposed heuristics.

\medskip\medskip

\noi \textbf{Keywords: Deep reinforcement learning,  Order batching, Sequential decision making, Machine learning, Warehousing, E-commerce   
}  

\end{abstract} 

\section{Introduction}
Warehouse management systems play a pivotal role in the supply chain strategy. Warehouse management systems mainly focus on storing and moving goods within a warehouse by performing several operations (including shipping, receiving, and picking). Order picking is the process of retrieving items from their locations in a warehouse \cite{petersen1997evaluation,petersen1999evaluation,de2007design}. This process is a significant operation in a warehouse and according to \cite{de2007design},  \cite{bartholdi2014warehouse}, and \cite{boysen2019warehousing}, the order picking cost are estimated to account for 55\% of the total warehouse operating cost. 

In the order picking process there are two basic order-picking decisions: picking-by-order and picking-by-batch. In the first one, the picker collects all the required items of one order in a pick tour. When picking-by-batch is performed, several orders are placed in a group (or a batch), and subsequently, the picker collects all the required items for that batch in a single pick tour. Choosing between these two decisions has its advantages and disadvantages. Picking-by-order provides short lead times but bears the risk of less productive walking distances, whereas batching has a longer lead time but increases the productivity of the picker. 

Besides the decision on whether picking-by-order or picking-by-batch should be applied for each arriving order, determining in which order the formed batches or orders should be collected such that orders are delivered before a due date (or cut-off time) is also a crucial problem to address in order picking. In a competitive e-commerce market, on time deliveries are of great importance to customer satisfaction. 

In this paper, we study an Order Batching and Sequencing Problem (OBSP) that determines how orders should be picked and sequenced such that the number of tardy orders is minimized and orders are shipped before their cut-off time. 
The existing work of the OBSP considers either a Person-to-Goods (PtG) or Goods-to-Person (GtP) storage system in the warehouse. 
In a PtG system, the picker collects orders following a certain pick route. If orders are small, the picker walks long unproductive distances between shelves and the central depot. However, if many orders are required, the throughput in a PtG system becomes a problem in meeting the cut-off times. The GtP systems apply automated solutions to bring the goods to a picker that is located at a workstation. The GtP provides fast order picking and also greatly reduces the need for operator labor. However, The GtP systems have a significantly higher investment level than the PtG system. In this paper, we consider a combination of PtG and GtP system. This hybrid storage system has been considered recently in practice. However, to the best of our knowledge, there is no existing solution that solves the order batching and sequencing problem for the combined PtG and GtP system.  


The order batching problem is known to be NP-hard if the number of orders per batch is larger than two \cite{gademann2005order}. The existing algorithms are therefore mainly constructive heuristics and metaheuristics. There is no existing approach to solve the OBSP that includes both picking-by-order and picking-by-batch decisions in a combined storage system. Moreover, due to the more complex problem setting, adapting the existing heuristics to our OBSP might not work well. 

In this paper we approach the OBSP using one of the popular machine learning approaches, called Deep Reinforcement Learning (DRL), which learns a good batching and sequencing strategy by interacting with the environment. DRL algorithms integrate Reinforcement Learning (RL) and Deep Neural Networks (DNN). DRL has been applied in many problem domains, from playing games \cite{mnih2013playing}, shunting trains \cite{peer2018shunting}, to scheduling jobs \cite{rummukainenpractical}.  
These existing DRL approaches have demonstrated the ability of DRL on handling uncertainties. In addition, because of the approximation power of DNN, DRL is able to capture a high dimensional state space, which is beneficial to the studied OBSP with the combination of the PtG and GtP system. 
In e-commerce and our case, uncertainties come from order uncertainty and seasonal peaks. With DRL, the learning agent might learn during which period of the day picking-by-order, picking-by-batch or the combination of the two is preferable, and what to do in case of a small disruption in peaks etc. A DRL agent is trained on sets of problem instances (or data), and once a good model is created, it can adapt to new situations. Therefore, the model learns to  respond to different situations in a fast and reliable way. This property of the approach is highly desirable in the current, high dynamic e-commerce setting. However, in contrast to heuristics or meta heuristics, DRL is much more complex to model and executed actions can be difficult to interpret.  Furthermore, the complexity of the environment also has a negative effect on the computation time to train the model. In this paper, we focus on the modelling and application of DRL in the studied order batching and sequencing problem. 

Our work in successfully creating a learning based order batching and sequencing algorithm  
contributes to both the fields of warehousing and to that of applying deep reinforcement learning to find strategies in this domain. We are the first to apply DRL to find strategies that solve the OBSP. Besides approaching the OBSP with DRL, we are also the first to apply DRL in the warehousing domain itself. We incorporate the combination two storage areas with two picking strategies: picking-by-order and picking-by-batch. Current applications mainly focus on solving the Order Batching Problem either for a PtG system or a GtP system and only few studies include picking-by-order. We apply an Actor-Critic method called the Proximal Policy Optimization algorithm to obtain a strategy for our problem. We  evaluate the DRL algorithm with several heuristics as a practically relevant benchmark, on the problem instances derived from real dataset. 

In the remainder of this paper, a brief overview of OBSP and DRL related literature is presented in Section \ref{Section: Literature review}. Subsequently, the dynamics of the OBSP and the simulation model that represent the warehouse are described in Section \ref{Section: Problem def}. In Section \ref{Section: DRL approach} the DRL model and environment are discussed and the Proximal Policy Optimization algorithm is explained. In Section \ref{Section: Experiment Setup}, the experiment evaluation of our work is described. Here, the performance of our DRL approach is compared to heuristics that are proposed by practitioners. In section \ref{Section: Conclusions} we elaborate on our conclusions.

\section{Related work}\label{Section: Literature review}
This section presents related work on Order Batching and Sequencing Problems (OBSP) in warehousing and Deep Reinforcement Learning (DRL) approaches to solving industrial optimization problems. 
We refer to \cite{boysen2019warehousing} and \cite{sutton1998introduction} for overviews of warehousing in e-commerce and reinforcement learning, respectively. 

\subsection{Order Batching and Sequencing}\label{Subsection:Literature OBSP}
The order batching problem is known to be $NP$-hard if the number of orders per batch is larger than two \cite{gademann2005order}. Due to its complexity, the existing algorithms for the OBSP are mainly heuristics: constructive solution approaches and metaheuristics. 
In this section, we will discuss the existing heuristics on OBSP where the objective is to minimize 
the tardiness. 

The author of \cite{elsayed1991} was one of the first that assumed distinct due-dates for orders for an OBSP, where seed and savings algorithms are developed to batch and sequence the orders to tours such that the total travel time and the total tardiness of retrievals per group of orders are minimized. In a subsequent work (\cite{elsayed1993sequencing}), the authors focus on minimizing the tardiness for an OBSP in a GtP system. They consider picking-by-batch and reduce the number of orders per batch towards one to analyze its affects. When the average number of orders per batch is close to one, the authors state that that batching does not provide significant improvements when compared with picking-by-order. 
In \cite{lee1995scheduling} the authors extend the problem of \cite{elsayed1993sequencing} with penalties for early retrieval and suggest a sequencing procedure. They show which of the 4 heuristics are superior to sequence the storage and retrieval requests to improve the due date related performance. 
The authors of \cite{elsayed1996order} also consider the OBSP for an GtP system, where they include the incoming but also the outgoing of goods(r-orders). Furthermore, they introduce and evaluate several heuristic rules for both picking-by-batch and picking-by-order in their problem. 
In \cite{tsai2008}, two algorithms based on Genetic Algorithm (GA) are developed to solve the OBSP in PtG system. One algorithm that finds the optimal batch picking plan by minimizing the sum of the travel cost, earliness and tardiness penalty. The second algorithm searches for an optimal travel path in a batch by minimizing the travel distance. 

In \cite{henn2013metaheuristics}, the authors state that since the number of possible batches grows exponentially with the number of orders, the use of heuristics for OBSP seems unavoidable. They propose an iterated local search (ILS) heuristic  
and an attribute-based hill climber (ABHC) heuristic based on a tabu search. 
The proposed approaches are compared with the Earliest Due Date (EDD) heuristic, and show an 46$\%$ improvement. 
Later in 
\cite{bustillo2015algorithm}, a Variable Neighborhood Search algorithm is proposed that exploits the idea of neighborhood change in a systematic way and outperforms the ILS proposed by \cite{henn2013metaheuristics}. 
The authors of \cite{menendez2017general} improve the total tardiness over that in \cite{henn2013metaheuristics} using 
a more aggressive improvement strategy. 
The proposed algorithm improves the current solution by selecting only those batches that contain at least one order with associated tardiness. For those batches, the procedure only focuses on orders without a tardiness value, since they could be retrieved later without affecting the objective function. Indeed, the move of these orders to a different batch could positively affect to those that currently have an associated tardiness.

\setlength{\tabcolsep}{3pt}
\begin{table}[ht]
\centering
\begin{tabular}{|l|l|l|l|l|l|}
\hline
& \textbf{System} &  \textbf{Algorithm}   &  \textbf{Orders}   & \begin{tabular}[t]{@{}l@{}}\textbf{Items/}\\\textbf{order}\end{tabular} & \textbf{Instances} \\ \hline

\cite{elsayed1991} & GtP  & Priority and Seed Rule &  8-12 & 35-65  & 30 \\ \hline
 \cite{elsayed1993sequencing}  & GtP& \begin{tabular}[c]{@{}l@{}}3 step priority index\\   algorithm\end{tabular}  &  20 & 35-65 & 24 \\\hline
\cite{lee1995scheduling} & GtP  & \begin{tabular}[c]{@{}l@{}}MTP, SPT, SPT-NL\\   \& LPT rules\end{tabular}    &  3-50 &  absent & 150 \\ \hline
\begin{tabular}[c]{@{}l@{}}\cite{elsayed1996order}\end{tabular} & GtP  & Priority and Seed rule   & 
8-12 &  35-65   & 30 \\ \hline
\begin{tabular}[c]{@{}l@{}}\cite{tsai2008}\end{tabular} & PtG  & Genetic Algorithm  &  25-250  & 30-400   & absent    \\ \hline
\begin{tabular}[c]{@{}l@{}}\cite{henn2013metaheuristics}\end{tabular}  & PtG                              & \begin{tabular}[c]{@{}l@{}}Iterated Local Search\\   \& Attribute-Based Hill \\ Climber\end{tabular} & 20-60 & 5-25 & 1600 \\ \hline
\cite{henn2015order} & PtG & \begin{tabular}[c]{@{}l@{}}Variable Neighborhood\\   Search and Descent\end{tabular} & \begin{tabular}[c]{@{}l@{}}100 or\\200\end{tabular} &  5-25    & 4800               \\ \hline
\begin{tabular}[c]{@{}l@{}}\cite{bustillo2015algorithm}\end{tabular}   & PtG                              & \begin{tabular}[c]{@{}l@{}}General Variable\\   Neighborhood Search\end{tabular}                     &  20-80 & 5-25 & 4800 \\ \hline
\begin{tabular}[c]{@{}l@{}}\cite{menendez2017general}\end{tabular} & PtG  & \begin{tabular}[c]{@{}l@{}}General Variable\\   Neighborhood Search\end{tabular}                     &  20-80   & 5-25 & 96 \\ \hline
\end{tabular}

\caption{Overview of the OBSP solutions with tardiness minimization.}
\label{table:Overview}
\end{table}

In Table \ref{table:Overview}, we provide an overview of the scope and approaches of the existing work for solving the OBSP. To summarize, all the discussed related work consider either a Person-to-Goods storage system or a Goods-to-Person storage system. To the best of our knowledge, there is no algorithms available that solves the OBSP with two types of storage systems. Moreover, although almost every study takes into account the capacity constraints of batches and pickers, few studies considers workstation constraints as we do in this paper. 

\subsection{Deep Reinforcement Learning for solving optimization problems}\label{Subsection: DRL literature}

\begin{figure}[ht]
\centering
  \includegraphics[width=0.8\linewidth]{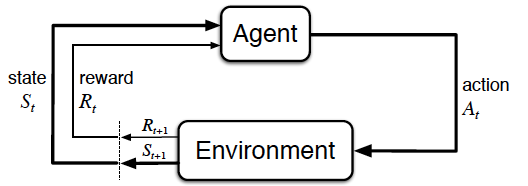}
  \caption{The agent–environment interaction in a Markov decision process, from \cite{sutton1998introduction}.}
  \label{fig:RL-agent}
\end{figure}

Recently, there is a great progress in developing machine learning (ML) methods to solve NP-hard problems. A popular line of the ML methods is \textit{Deep Reinforcement Learning (DRL)}, which is the integration of Reinforcement Learning (RL) and Deep Neural Networks (DNN).   
\paragraph{Reinforcement Learning} 
In an RL approach, an agent interacts with an environment (see Figure \ref{fig:RL-agent}). At each time step, the agent observes the current state $S_t$ of the environment, and selects an action $A_t$ to perform. Following the action, the agent receives a reward $R_t$ and the environment 
moves to a new state $S_{t+1}$. 
The state transitions are assumed to satisfy the Markov property, that is, the state transition probabilities depend only on the state $S_t$ and the action $A_t$ taken by the agent, independent of all previous states and actions. 
The agent has no prior knowledge of the environment in terms of state transitions or what the reward is. By interacting with the environment, the agent can observe such knowledge. The learning goal of the agent is to maximize the expected cumulative reward over the relevant time horizon. 
For more details of the RL, we refer to \cite{sutton1998introduction}. 

The agent chooses a certain action based on a policy, which is a probability distribution over actions in states: $\pi: \pi(s,a)\rightarrow [0,1]$. Since there are many possible states and actions in most real-world problems, a function approximator is often used to enable generalization from seen states to unseen states. Deep Neural Networks have been successfully used as function approximators for solving large-scale optimization tasks.

\paragraph{DRL applications} The Traveling Salesman Problem (TSP) and closely related Vehicle routing problem (VRP) have been become popular problems that are approached using DRL in the AI community.
The authors of \cite{bello2017neural} use policy gradient and a variant of Asynchronous Advantage Actor-Critic (A3C) algorithm  \cite{mnih2016asynchronous} to train a DNN. 
In \cite{nazari2018reinforcement} the authors study the VPR with the objective of minimizing the total route length while satisfying the demand from all customers. They employ the Asynchronous Actor-Critic Agents (A3C) algorithm \cite{mnih2016asynchronous}. 
The authors of \cite{kool2018attention} investigate a TSP where the pointer network is incorporated with attention layers and they train the model with the REINFORCE algorithm. The authors state that the operational constraints in the problem often lead to many variants of combinatorial optimization problems for which no good heuristics are available. This is the case for the OBSP. In \cite{costa2020learning}, the authors propose a deep reinforcement learning algorithm trained via Policy Gradient to learn improvement heuristics based on 2-opt moves for the TSP.


In supply chain management, the authors of \cite{oroojlooyjadid2017deep} examine the applicability
of DRL to the beer game, 
which is a simplified model of a serial supply chain with four echelons and is played to demonstrate the bullwhip effect in supply chains. The Deep Q-learning (DQN)  
algorithm \cite{mnih2015human} is used in their approach. 
in \cite{gijsbrechts2018can}, the author solves a dual sourcing inventory problem wherein 
an inventory can be replenished from a fast but expensive source or from a regular, cheaper source with longer lead time. The authors show how
the A3C algorithm can be trained to produce policies that match performance 
of several existing approaches and conclude that DRL provides solid inventory policies for environments for which no heuristics have been designed, and can inspire new policy insights.

In manufacturing systems, the authors of  \cite{waschneck2018deep} simulate an DRL approach in a dynamic manufacturing environment to allocate waiting jobs to available machines. They apply the DQN algorithm and have an agent for each work center. 
In \cite{rummukainenpractical}  a job scheduling for production control environment is considered. The authors model their system as a Semi-Markov decision process (SMDP) and interact with a simulation model. By applying the proximal policy optimization (PPO) algorithm \cite{DBLP:journals/corr/SchulmanWDRK17}, the authors shows good performance of the DRL approach in the  experimental application. They also state that DRL is best suited to applications where no existing control methods are satisfying, which is also the case with our OBSP problem.

\subsection{Summary}

We first investigate the existing algorithms for solving the OBSP. We show that although many heuristics already have been developed, they are problem-specific.  Moreover, the problem we consider, a combination of a PtG and GtP storage systems, has not been investigated in the literature. Hence, no existing algorithms can be directly applied to solve our OBSP. In the experiment section, we will fine-tune some existing heuristics to compare the performance of our DRL approach. Furthermore, due to its complexity and strategies that are required to be learned from the agent, reinforcement learning seems very well suitable for this problem.

Deep learning makes Reinforcement Learning applicable to more complex environments such as our OBSP. More and more research have shown the effectiveness of applying DRL in the complex decision and optimization problems. To the best of our knowledge, no DRL based approach exists yet that tackles order batching and sequencing problems. 
Among different DRL algorithms, the value method such as the DQN is powerful in learning difficult strategies as it considers each state-action pair. However, when action spaces and observation space become large, the required computational power for the DQN can become a challenge. Policy search methods such as PPO require less computation power since only the policy is updated and not individual state and action pairs. Inspired by the work of \cite{rummukainenpractical}, we apply PPO to solve the OBSP. 

\section{The warehouse setting 
and problem definition}\label{Section: Problem def}

\subsection{The warehouse setting with two picking decisions}
\begin{figure*}[h]
  \includegraphics[width=\linewidth]{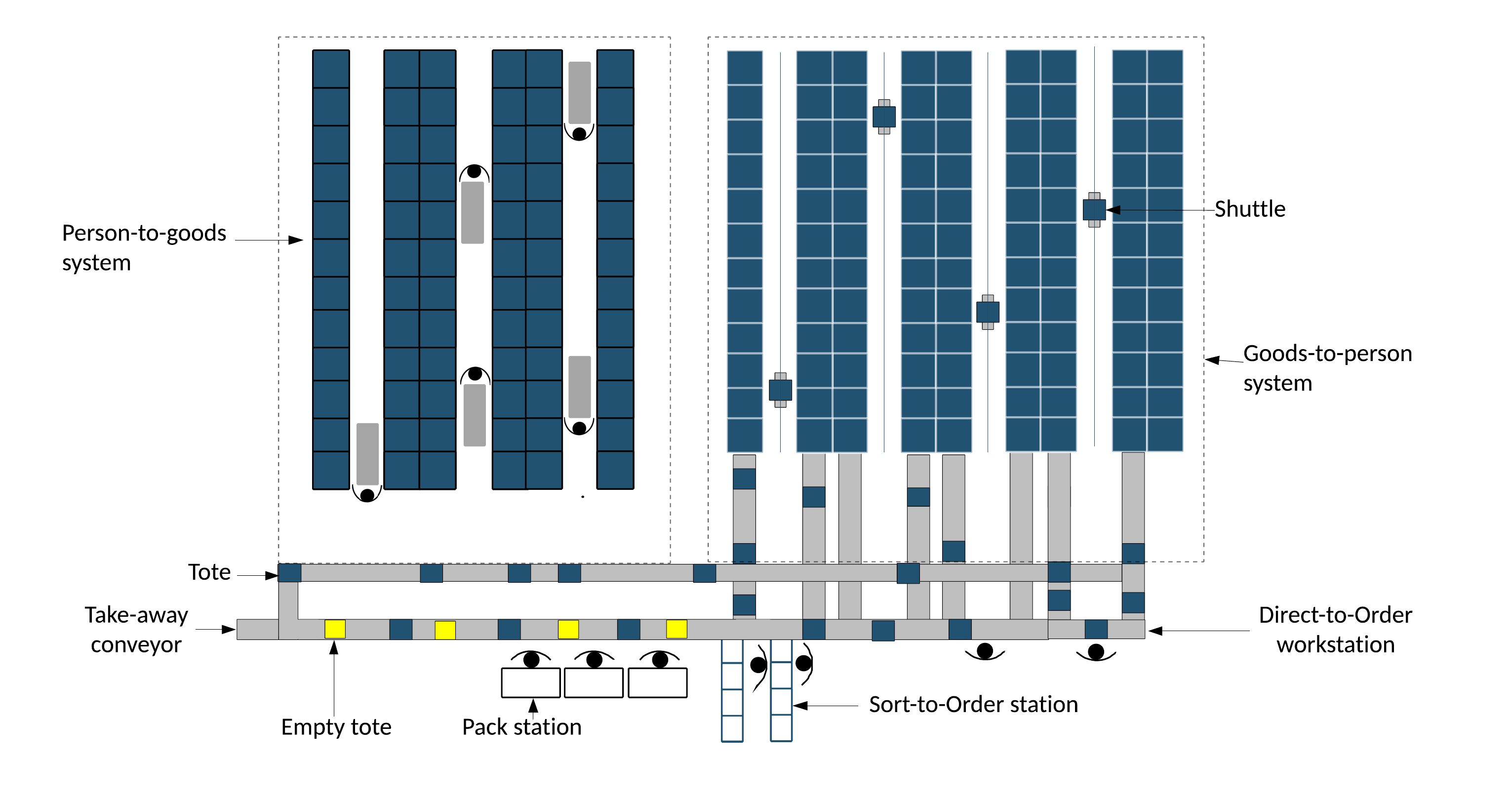}
  \caption{Person-to-Goods and Goods-to-Person order picking system with a pack station, sort-to-order and direct-to-order workstations.}
  \label{fig:warehousesetting}
\end{figure*}

We consider a warehouse setting with a PtG and a GtP storage system to store SKU and consider three different type of workstations to consolidate the orders, see Figure \ref{fig:warehousesetting} for an overview.

In the PtG storage area, a person (picker) has to walk through the warehouse to pick the goods. This system is constrained by the number of pickers that can collect items and the number of items that can be included on the picking cart. Each picker can collect 50 items on a picking cart. 
In the GtP goods are automatically transported to people at picking stations in unit loads (totes or bins). The GtP is constrained by the number of shuttles. Shuttles are automated vehicles that bring totes with items to one of the lifts of the GtP system. Lifts are connected to a conveyor of one of the workstations. Shuttles have the advantage to travel through every aisle at a relatively high speed. Therefore, the vehicles offer much higher retrieval capacity and are also significantly more flexible
in capacity compared to a crane-based AS/RS that is mostly fixed within one aisle \cite{azadeh2017robotized}.

For both storage systems, orders can be picked-by-batch or picked-by-order and be released to the workstations once there is sufficient capacity available. Depending on the picking decision and storage system, a workstation is required to consolidate the order. Totes or bins that contain items or orders can be released to three different types of workstations: \textit{Direct-to-Order (DtO) workstations}, \textit{Sort-to-Order (StO) workstations}, and \textit{pack stations}. The number of workstations of each type is constrained.

At a \textit{DtO workstation} a picker removes items from a product tote and collects them into a carton box. At this workstation, only one order at a time is packed and sent for shipping, thus this workstation is used for picking-by-order. If an order consists of multiple order lines (thus multiple different SKUs), multiple product totes are provided. The product totes arrive in sequence and once the each item is picked from product tote, it is placed in the carton box.

An \textit{StO workstation} can be used for picking-by-order and picking-by-batch and is divided into three processes: \textit{sorting}, \textit{buffering}, and \textit{packing}. During \textit{sorting}, a picker removes the items from a batch tote and sorts them into a put wall. A put wall is a simple rack with shelves separated into multiple locations with a maximum of approximately 50 order locations. The locations are temporarily assigned to a unique order. Once the put wall is filled with all the items for each order, the put wall is placed in the \textit{buffering} area. Here, it waits until an operator is available to pack all the orders. At \textit{packing}, the operator first requests a put wall and drives the remote put wall to the packing area. The operator then places all the items of each order into a carton box and sends the carton box via a takeaway conveyor to shipping.

The \textit{pack station} is the simplest workstation of the three and is only used for the picking-by-batch decision. These batch totes do not require sorting because each item in the batch is one order. The orders packed at this station thus only contain one SKU.

\begin{figure}[h]
 \centering
  \includegraphics[width=85mm,scale=1.5]{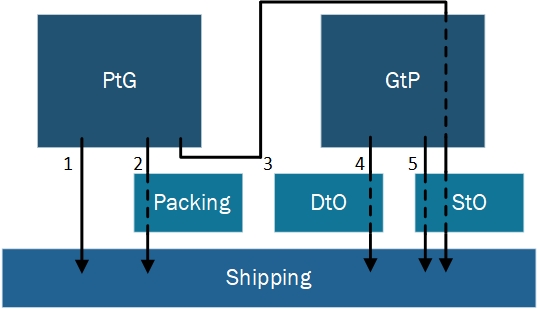}
  \caption{Picking routes of the warehouse setting}
  \label{fig:Pickingroutes}
\end{figure}

With the two storage systems and the three types of workstations, 5 picking routes are available. Figure \ref{fig:Pickingroutes} shows a schematic overview of these five routes through the warehouse.

In total there are ten ways to pick orders (see Table \ref{table:Picking methods with routes}). Besides the picking decision and the storage location, the order type also has influence on which workstation is required to consolidate the order. As mentioned earlier, orders can consist of one or more order lines. The number of order lines classify an order either as a Single Item Order (SIO) or a Multiple Item Order (MIO). An MIO containsmultiple items of multiple SKU. In a SIO, only one item and thus one SKU is ordered. Whether an order is a MIO or a SIO will have a strong influence on the further processes. 

\begin{table}[ht]
\begin{tabular}{|l|l|l|l|l|}
 \hline
 \textbf{Order with}& \textbf{SKU(s) stored in }&\textbf{Picking decision}&\textbf{Workstation}&\textbf{Route}\\
 \hline
 
 SIO   & PtG   &Pick-by-Batch&   Pack station& 2\\
 \hline
 SIO   & GtP   &Pick-by-Batch&   Direct-to-order& 5\\
 \hline
 SIO   & PtG   &Pick-by-Order&   - & 1\\
 \hline
 SIO   & GtP &Pick-by-Order &   Direct-to-order& 4\\
 \hline
 MIO    &PtG &Pick-by-Batch&  Sort-to-Order & 3 \\
 \hline
 MIO &   GtP  &Pick-by-Batch& Sort-to-Order& 5 \\
 \hline
 MIO   & PtG   &Pick-by-Order &  - & 1\\
 \hline 
 MIO   & GtP &Pick-by-Order &   Direct-to-order& 4\\
 \hline
 MIO&PtG and GtP & Pick-by-Batch&  Sort-to-Order& 3\\
 \hline
 MIO &  PtG and GtP  & Pick-by-Order & Sort-to-Order& 3\\
 \hline
\end{tabular}
\caption{Picking methods and routes for the warehouse setting}
\label{table:Picking methods with routes}
\end{table}

To model our system we make a number of assumptions and simplifications. First, we assume that each SKU is either stored in the PtG or in the GtP system and not in both. Second, we do not consider the exact storage location of an SKU. Instead, we simulate the transportation time for each SKU. Third, we assume that transporting totes between processes is always possible; possible deadlocks or queues on these conveyors are not taken into account. Fourth, replenishment of SKUs and transportation of empty totes is left out of scope.


\subsection{Problem formulation}

The OBSP problem that is studied in this paper can be formulated as follows. Given a warehouse setting with pickers, shuttles, packing, StO and DtO workstations and order arrivals events, the proposed algorithm needs to sequentially (1) assign orders to pickers and shuttles, and (2) assign picked orders to workstations, such that the capacity constraints of the pickers, shuttles, packing, StO and DtO workstations are not violated, and the number of orders that are shipped on-time is maximized. 
 
Let us consider an example of this problem in which the DRL agent needs to determine for 400 orders when and how to pick them. Approximately 75\% of these 400 orders are SIO orders and approximately 25\% are MIO orders and both are stored in either the PtG or the GtP system. All orders have a random cut off time and have to be shipped within 15 to 45 minutes. Preferring one picking method over the other may cause some orders to be shipped on time and others to become tardy. Furthermore, sacrificing one critical order and let it become late may result in other orders that can then be shipped on-time. Therefore there are many ways and when orders can be picked and consolidated.

\section{Deep reinforcement learning approach}\label{Section: DRL approach}
We solve the OBSP using deep reinforcement learning, by modeling it as a Semi-Markov Decision Processes (SMDP). In the SMDP the state of the system is primarily composed of the orders of each type that must be picked and the current capacity that is available for picking those orders. The actions of the system are the decisions to pick a particular order in a particular way. When an action is taken, this results in a new state and a reward for taking the action. The reward is primarily determined by the number of orders that is picked on time. The DRL agent learns from those rewards what the optimal picking action is in a particular state of the warehouse.

The new state and reward are computed using a simulation model of the warehouse. In that sense the simulation model forms the environment of the DRL agent. The simulation model computes the state and reward that result from a particular choice of action of the DRL agent, by simulating the effect of that action on the state of the warehouse, the duration of the picking action and, consequently, the orders that are not picked on time.

In the remainder of this section, we present the SMDP that models our OBSP (see Section~\ref{Subsection:SMDP}). We then present the simulation model that simulates the warehouse (see Section~\ref{Subsection:Environment}) and finally the algorithm (see Section~\ref{Subsection:Algorithm}) that the DRL agent uses to learn which actions are optimal in a particular state.

\subsection{SMDP formulation of the order batching and sequencing problem}\label{Subsection:SMDP}
In order to create model in which the DRL agent can train and learn a strategy, we define a SMDP which consists of four components: (1) the time to transition  \textbf{$\tau$}; (2) the set of states \textbf{$S$}; (3) the set of actions \textbf{$A$}; and (4) a reward function \textbf{$R$}. Another main component of an SMDP is \textbf{$P$}, the transition probabilities, which represent the uncertainty about what the next state will be. In our system this uncertainty is the result of not knowing which capacities become free and when orders arrive into the system. In DRL these transition probabilities are learned by the DRL agent. 

\subsubsection{Time to transition $\tau$}
Since our environment heavily depends on time, a SMDP is used instead of a Markov Decision Proces (MDP). 
In an MDP the time to transition \textbf{$\tau$} is fixed and in each state, the agent can choose a picking action or to ``do nothing'', after which it waits for a time \textbf{$\tau$} to take the next decision. In that case we run the risk of setting \textbf{$\tau$} too low, thus increasing the number of ``do nothing'' actions, because it does not allow for enough time for the environment to change state. We also run the risk of setting \textbf{$\tau$} too high, possibly leading to cases in which the environment changes state before the agent is ready to take the next decision. As a result, the environment remains idle until a new action is performed. This is undesirable because it may negatively effect the throughput time of orders, which are waiting idly for a decision to be taken by the agent. This, in turn, negatively affects the number of orders that are delivered on-time and consequently the reward function.

In comparison, in an SMDP, the time to transition \textbf{$\tau$} is not fixed. In our system, it is determined by the simulator: the time to transition is the time until the next order arrives or the next order is picked. As a consequence, the DRL agent is always able to perform an action and there are no unnecessary idle times. This approach is similar to the approach taken by \cite{rummukainenpractical}, who also make use of a simulation model that contains capacity that can become idle.

\subsubsection{State Space \textbf{$S$}}
To model a state in such a way that it captures all the relevant information of our warehouse setting, we define three main components: (1) Current remaining orders for picking; (2) Capacity availability in the warehouse system; and (3) Extra information beneficial for learning, i.e., number of tardy orders, number of processed orders and current simulation time. We now describe these components in detail.

\textbf{Current remaining orders for picking} keeps track of the number of orders 
$\mathnormal{O_{c_il_je_k}}$
for each composition 
$c_i\in \{c_1,c_2\}$, 
storage location 
$\mathnormal{l_j\in {l_1,l_2,l_3}}$, 
and earliness/tardiness 
$\mathnormal{e_k\in \{e_1,e_2,e_3\}}$, where 
\begin{enumerate}
    \item we consider two compositions: $\mathnormal{c}_1$ denotes a SIO order and $\mathnormal{c}_2$ a MIO order; 
    \item there are types of storage locations considered: a PtG area $\mathnormal{l}_1$ and a GtP are $\mathnormal{l}_2$. Since SKUs of a MIO can be stored in both locations, we define the third location $\mathnormal{l}_3$;  
\item orders are divided into three earliness/tardiness categories.  $\mathnormal{e}_3$ represents the orders that have to be shipped within 15 minutes, $\mathnormal{e}_2$ represent the orders that have to be delivered within 40 minutes and $\mathnormal{e}_1$ include all the orders that have to be shipped in 41 minutes or later. 
\end{enumerate}

With two compositions of order types, three storage possibilities and three earliness/tardiness categories, 15 different type of $\mathnormal{O_{c_il_je_k}}$ exist in the first part of our state. 

\textbf{Capacity availability in the warehouse system} is also included in the state as the second part of the state representation. All capacities are represented by a pipeline variable that includes the current available capacity plus a (virtual) queue variable. This queue variable is included to make sure that the resource does not become idle too quickly between states. In case no queue is included and capacity becomes available, the agent could choose to process an order that requires the capacity that has just become available. However, if the agent chooses a different order that requires different capacity, the capacity that has just turned available stays available and the resource thus stays idle. If a queue variable is included, more orders are put into the system and end up in a queue. When orders are waiting in the queue and the resource capacity becomes available, the resource can start directly processing these orders from the queue without having to wait for the agent.

The pipeline variable $\mathnormal{p}=r+R$ represents the capacity availability of the PtG area in our state, where 
$\mathnormal{r}$ is the current number of available pickers and $\mathnormal{R}$ is a constant variable, representing the queue length of the PtG area. The length of the queue $\mathnormal{R}$ is equal to the maximum number pickers available to ensure that when pickers become available, sufficient orders are available for processing. 

The pipeline variable $\mathnormal{g}=\mathnormal{h} + \mathnormal{H}$ represents the capacity availability of the GtP area in our state, where 
$\mathnormal{h}$ denotes the current number of available shuttles and $\mathnormal{H}$ is a constant variable  representing the queue length of the GtP area. Picked batches in the PtG area, are placed in the GtP queue. This increases the average occupation of the GtP queue. However, to ensure that there are always enough orders/batches in the queue, we also include a constant variable $\mathnormal{H}$. This is especially advantageous at the beginning of the episode since processing batches can take some time before being available to the shuttles. 

The pipeline variables  $\mathnormal{d}$, $\mathnormal{v}$ and  $\mathnormal{b}$ represent the capacity availability's of the DtO workstation, StO workstation and packing station respectively. At the workstations, capacities are expressed in the number of totes that can be placed in the queue plus a virtual queue. Orders that require capacities at the workstations have to be picked first. Therefore, the queue length is only increased once orders arrive in the queue. The maximum queue length of each workstation is based on the tote processing time at the workstation and the time it takes to place a tote in the queue. However, if no virtual queue variable is included in the state and only the queue size, slots of the queue are reserved for that specific tote and if all slots are reserved, no more totes can be placed. As a result of the picking time, it will be the case that the workstation is then waiting for the totes to arrive. When including a virtual queue, more orders can be released into the system such that the throughput is increased. 

Therefore, the second part of the state representation regarding the capacities is represented by the following variables:  $(\mathnormal{p, g, d, v, b})$.

As \textbf{extra information beneficial for learning}, we include the variables $\mathnormal{t, n}$ and $u$ in the last part of the state representation. Variable $\mathnormal{t}$ represents the number of tardy orders, $\mathnormal{n}$ represents the number of processed orders and $\mathnormal{u}$ represents the current simulation time. An order becomes tardy when the order has not been shipped before its cut-off time and is represented by the variable $\mathnormal{t}$. Additionally, the number of processed orders $\mathnormal{n}$ and the current simulation time $\mathnormal{u}$ are represented in the state. Both increase as we get closer to the end of the episode. Variables $\mathnormal{t, n}$ and $u$ are updated once the state changes in capacity or order arrivals, making sure that the state only transits to the next state based on changes in capacities or order arrivals. 

Note that many other components can be included in the state. Components such as the number of items of each order, processing times of next order/batches or the number of orders waiting in the queues could also be included in the state representation. However, there is a challenge in representing these variables. Including information such as processing times or the number of items for each individual order will increase the size of the state vector. As a result,  the number of possible actions is also increased and consequently more computation time is required.

Our state is represented as a one-dimensional vector containing all the different types of orders, capacities and extra information beneficial to learning. In total our state space vector consists of 23 variables as follows:

$\big(\mathnormal{O_{c_1l_1e_1}, O_{c_1l_1e_2}, O_{c_1l_1e_3}, O_{c_1l_2e_1}, O_{c_1l_2e_2}, O_{c_1l_2e_3}, O_{c_2l_1e_1}, O_{c_2l_1e_2}, O_{c_2l_1e_3},}$

$\mathnormal{O_{c_2l_2e_1}, O_{c_2l_2e_2}, O_{c_2l_2e_3}, O_{c_2l_2e_1}, O_{c_2l_2e_2}, O_{c_2l_2e_3}, p, g, d, v, b, t, u, n}\big) $

As an example, the vector at the start of an episode could look like:  
\begin{center}
$\big(\mathnormal{180, 12, 0, 55, 10, 0, 38, 5, 0, 25, 0, 0, 5, 0, 0, 10, 12, 50, 75, 25, 0, 0, 0 }\big)$
\end{center}

In total there are 330 orders (180 + 12 + 55 + 10 + 38 + 5 + 25 + 5 = 330) in this state vector of which $12 + 10 + 5 = 37$ have to be shipped within 40 minutes (belong to the $\mathnormal{e_2}$ earliness category). The other orders have to be shipped within 41 minutes or more (belong to the $\mathnormal{e_1}$ earliness category). In this example the picker pipeline variable $\mathnormal{p}$ is ten and the shuttle pipeline variable $\mathnormal{g}$ is 12. Furthermore, at the DtO, StO and packing workstations, 50, 75 and 25 totes can be placed in the virtual queue respectively. 

Depending on the picking decision and the type of order, capacities are required for each action. For example, when choosing to process one order of the first index in the vector; $O_{c_1l_1e_1}$, only one picker is required. In the next state the $O_{c_1l_1e_1}$ variable will represent 179 orders instead of 180 and $g$ will represent nine pickers instead of ten. When performing a batch action for $O_{c_1l_1e_1}$, one picker and one tote at the packing station is required. When performing this batch action, $O_{c_1l_1e_1}$ in the next state is reduced from 179 orders to 169 orders. the next state.  $g$ and $b$ are both reduced by 1. After performing these two actions, the agent will end up in the following state.
\begin{center}
$\big(\mathnormal{169, 12, 0, 55, 10, 0, 38, 5, 0, 25, 0, 0, 5, 0, 0, 8, 12, 50, 75, 24, 0, 0.001, 0 }\big)$
\end{center}
 
Note that the current simulation time $u$ has also increased. When assigning orders to pickers or shuttles, the state changes. Assigning these orders takes a really small fraction of time to change the state. After this change, the agent is directly provided with a new state. 

Since the number of states is large, applying some state reduction methods to improve the learning speed is beneficial. In  \cite{afshar20state}, the authors use a tabular reinforcement learning algorithm to find the best aggregation strategy for reducing the state space. \cite{zhang2012minimizing} study a dynamic machine scheduling problem with a mean weighted tardiness objective function, where the authors  model their state with jobs by indicating if there are one or more jobs in the queue. 
In this paper, we use the idea similar to   \cite{zhang2012minimizing}. We introduce a maximum state representation constant $\mathnormal{M}$, and then all variables in the states are capped at this value. 
As the learning agent can choose to process a batch of maximum 10 to 12 orders in our setting, we set $\mathnormal{M}$ to 25. As an example, the state vector from the previous example is represented as the following:
\begin{center}
$\big(\mathnormal{25, 12, 0, 25, 10, 0, 25, 5, 0, 25, 0, 0, 5, 0, 0, 10, 12, 25, 25, 25, 0, 0, 0 }\big)$
\end{center}

However, by introducing the constant $M$, the variable $p$ and $g$ can be less than $M$. To increase the importance of $p$ and $g$ in the state vector, $p$ and $g$ are de-normalized to the maximum state representation constant. In other words, when $p$ and $g$ have maximum capacity available, $p$ and $g$ are equal to $M$. As an example, if the maximum state representation constant $M$ is set to 25 and $p = 10$, $p$ in the state is de-normalized to 25 such that $p = 25$ is represented in the state. This is applied for the shuttles and the pickers capacity variables as they can become relatively small numbers when orders are being processed. Processing orders is one of the main goals of the algorithm. As a result, $p$ and $g$ are desired to be fully utilized and therefore only indicating small values when becoming available. By de-normalizing $p$ and $g$, their importance is encouraged. 

We also de-normalize the variables $t$, $u$ and $n$ as these variables increase over time and can become greater than $M$. As an example, the initial state after de-normalizing becomes: 

\begin{center}
$\big(\mathnormal{25, 12, 0, 25, 10, 0, 25, 5, 0, 25, 0, 0, 5, 0, 0, 25, 25, 25, 25, 25, 0, 0, 0 }\big)$
\end{center}

Note that the difference with the previous example in which the vector is capped to a maximum constant $M$ of 25, in this example the pickers and shuttles are maximized such that recognizing these capacity availabilities is encouraged.

\subsubsection{Action Space \textbf{$A$}}\label{subsubsection:action_space}
We formulate an action space that directly maps to the orders available in the state. In total there are 15 different types of orders $\mathnormal{O_{c_il_je_k}}$ that can all be picked-by-order and picked-by-batch. Therefore, for each type of $\mathnormal{O_{c_il_je_k}}$ there are two possible actions resulting in 30 different actions. We also include a wait action that allows the environment to wait until capacity becomes available. This action must be added because no pick actions can be taken when no capacity is available for picking. Consequently, in total there are 31 actions available in our action space. 

When a picking-by-batch action is performed, a batch size of ten orders is created. The sequence of the orders and batches for the picking-by-batch and picking-by-order actions, is based on the tardiness; orders that have the lowest tardiness are picked first. 

Depending on the available orders $\mathnormal{O_{c_il_je_k}}$ in the state and on the capacities $(\mathnormal{p, g, d, v, b})$, the agent can choose to process a certain order. Choosing an order for which no capacity is available, or selecting an action for which no order is available, is considered an infeasible action. The wait action is only feasible when no capacities are available. In other words, if the wait action is chosen while there are capacities available to process at least one of the available orders, the wait action is then considered as an infeasible action.
When a feasible action is chosen, the order with the earliest cut-off time is processed. When an infeasible action is performed by the agent, the state does not change, no orders are processed and the agent can immediately take another decision. In addition, an infeasible action will also lead to a penalty, which is discussed in more detail in the next section. 

\subsubsection{Reward function \textbf{$R$}}
We construct a reward function $r(s, a, s')$ in a way that it provides a relatively high reward at the end of the episode and small penalties during the episode. We include penalties for orders that become tardy and for infeasible actions in the reward function as follows. 
\begin{equation}
  \textit{  r(s, a, s')} =
    \begin{cases}
      -0.005 & \text{if infeasible action} \\
      -0.0075 & \text{if tardy order} \\ 
      (1- w/N )^2 & \text{if episode terminates} \\
      0 & \text{otherwise}
      
    \end{cases}
    \label{Equation: Reward function}
\end{equation}

In the constructed reward function, $\mathnormal{w}=\mathnormal{t}+m$, where  $\mathnormal{t}$ denotes the number of tardy orders and $m$ the number of orders that have not been processed at the end of an episode. 
$m$ is greater than zero when an episode terminates too early. This can happen when too many orders have become tardy. $m$ then indicates the number of orders that remained in the state and have not been processed. If an episodes terminates, we determine the ratio of processed orders by dividing $\mathnormal{w}$ by the sum of the the total orders $\mathnormal{N}$ and subtracting it from 1. The end reward shows, therefore, the ratio of processed orders that have been shipped before its cut-off time and the reward increases exponentially as this ratio increases. As a result, the importance of fewer tardy orders to the agent is also increased. 

The penalty for infeasible actions (i.e. $0.005$) is relatively small compared to the end reward such that the end reward has the upper hand and influences the learning process the most. However, it is big enough such that the agent will less likely choose this action again when the same state is encountered. Making the penalty for infeasible actions too large, will result in a strategy in which the agent performs as little actions as possible to complete the episode such that the total penalties of infeasible actions are minimized. In this case, the penalty for infeasible actions influences the learning more than that for tardy orders (i.e. $0.0075$).

\subsection{Interaction with the environment: the simulation model}\label{Subsection:Environment}
\begin{figure}[ht]
 \centering
  \includegraphics[width=120mm,scale=1]{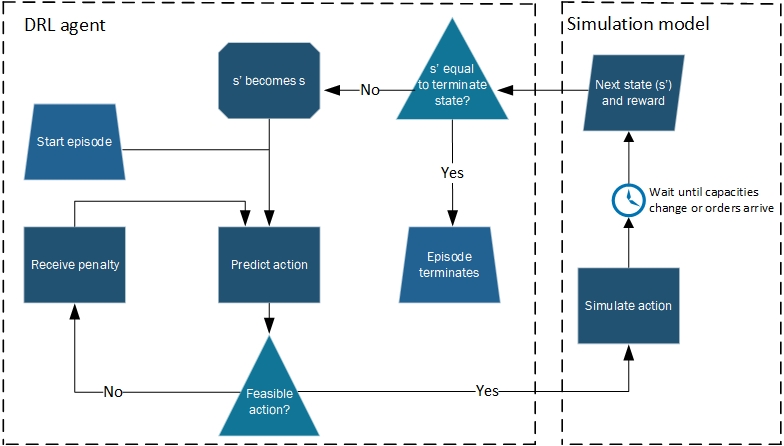}
  \caption{Flowchart of interaction between the DRL agent and the simulation model}
  \label{fig:Flowchart_interaction}
\end{figure}

The DRL agent learns which action is the best action in a particular state, by trying actions and observing how the environment responds to those actions (see Figure \ref{fig:RL-agent}). 
In our approach, the environment is formed by a simulation model that simulates how a picking action, with random arrivals of orders, leads to a new state of the warehouse and a reward for the agent. Figure \ref{fig:Flowchart_interaction} shows the flowchart of the interaction between the DRL agent and the simulation model.

At the start of an episode, given state $s$, the agent predicts the best action to perform.  
Next the agent determines if the predicted action is a feasible action. 
If the agent selects an infeasible action, the state does not change and the agent receives a penalty (see Equation \ref{Equation: Reward function}). After failing to select a feasible action, the agent can try again to select an action until a feasible action is chosen. These infeasible actions can be performed without interacting with the simulation model.

If the agent selects a feasible action, it passes on this action to the environment: the simulation model. The simulation model receives the action $a$ and simulates it until state $s$ changes to $s'$. The state changes when capacities change or when an order arrives into the system. Feedback is provided to the agent when orders have become tardy. This is determined once the order has been consolidated at one of the workstations and is ready for shipping. While the agent waits for capacities to become available or orders to arrive, time $\tau$ increases and tardy orders may leave the system. When all capacities are available and the agent selects to process an order, the state directly changes as capacities are diminished. The time to transition $\tau$ between these two states is almost zero. When an order is received by the simulation model, the model assigns a picker or shuttle to this order. Therefore, either $p$ or $g$ changes and feedback is directly provided back to the DRL agent and the simulation model is frozen until it receives another action. Whereas, $\tau$ is very small between states in which capacities are available. $\tau$ is larger when the wait action is chosen. In these situations, the duration of $\tau$ depends on arriving orders or capacity becoming available.

When the state changes, the new state is passed on to the DRL agent, which checks if the state represents the end of an episode. For our system, the episodes ends either when all orders have been picked or when the simulation time has been reached. The orders that have not been picked during the time frame are considered to be tardy orders. If the end of the episode is not reached, the flow repeats for the new state $s'$.

\subsection{Learning algorithm for DRL agent}\label{Subsection:Algorithm}

We use a Deep Neural Network as a function approximator in the DRL agent.  
The input layer of the DNN is the state representation of the given MDP. 
The output layer of the DNN represents the action space, i.e. the set of actions the DRL agent can choose from. 
The received rewards or penalties by the agent for different actions are used to update the parameters $\theta$ in the Neural Network. 
These parameters are called poicy parameters, and the policy can be then denoted by $\pi_\theta (a\ |\ s)$. 

We use the proximal policy optimization (PPO) algorithm  \cite{DBLP:journals/corr/SchulmanWDRK17}, which is a policy based algorithm that learns by performing gradient descent on the policy parameters $\theta$. 
 
The policy gradient methods seek the optimal approximation parameters online, which means that the algorithm does not apply a replay buffer to store past experiences and instead learns directly from the experiences that are encountered by the agent. Once a batch of experiences has been used to do a gradient descent update, the experiences are then discarded and the updated policy moves on to perform new actions.  
Policy gradient methods work by computing an estimator of the policy gradient and plugging it into a stochastic gradient ascent algorithm. The most commonly used gradient estimator has the form: 
\begin{equation}
    \hat{g} = \mathbb{\hat{E}}_t \Big[ \nabla_\theta\ \text{log}\ \pi_\theta (a_t\ |\ s_t) \hat{A}_t) \Big]
\end{equation}
where $\pi_\theta (a_t\ |\ s_t)$ is a stochastic policy and $\hat{A}_t$ is an estimator of the advantage function at time step $t$, which we will explain later. 
$\mathbb{\hat{E}}_t$ indicates the average over a finite batch samples, which is used to alternate between sampling and optimization. When implementing this in the neural network such that the objective function is the policy gradient descent, the estimator of $\hat{g}$ can be obtained by differentiating the following loss: 
\begin{equation}
    L^{PG} (\theta) =  \mathbb{\hat{E}}_t \Big[ \text{log}\ \pi_\theta (a_t\ |\ s_t) \hat{A}_t) \Big]
\end{equation}
where $\pi_\theta$ is our policy in which the neural network takes the observed state as an input and suggests the action to take as an output, indicated as log probabilities. The second term is the advantage function $\hat{A}_t$ which tries to estimate what the relative value is from the selected action $a_t$ in the current state $s_t$. $\hat{A}_t$ can be computed by subtracting the baseline estimate $V_t$ from the cumulative discounted reward $G_t$. $G_t$ 
is the weighted sum of all the rewards the agent received during each time step in the current episode, computed as follows: 
\begin{equation}
    G_t =  R_{t} + \gamma R_{t+1} + ... = \displaystyle\sum_{k=0}^{\infty} \gamma^k R_{t + k}
\end{equation}
where the discount factor $\gamma \in [0,1]$ is usually a value between 0.9 and 0.99, presenting the value of future rewards, i.e., a reward received k time steps in the future is worth only $\gamma^{k-1}$ times what it would be worth if it were received immediately. 
The advantage $\hat{A}_t$ is calculated after the episode sequence is collected from the environment, i.e., after all the rewards are known. The second part of the advantage function is the baseline estimate or the value function $V_t$. $V_t$ estimates the discounted sum of rewards given $s_t$. In other words, the value function attempts to predict what the final reward is going to be for this episode when in state $s_t$. Note that this value estimate is output of the neural network. 
The advantage function determines how much better was the action $a_t$ that the agent took, compared to the expectation for $s_t$.
Finally, by multiplying $\hat{A}_t$ with the log probabilities of the policy actions, the final optimization objective function is derived that is used in basic policy gradient methods. 

The PPO algorithm introduces Trust Region Policy Optimization (TRPO) to improve training stability. 
TRPO avoids parameter updates that change the old policy too much at one step. 
In TRPO, the objective function is maximized subject to a constraint on the size of the policy update. Specifically, 
\begin{equation}
    \max\limits_{\theta}\  \mathbb{\hat{E}}_t \Bigg[ \frac{\pi_\theta (a_t\ |\ s_t)}{\pi_{\theta\text{old}} (a_t\ |\ s_t) } \hat{A}_t  \Bigg] \\ \text{, subject to } KL_{\pi_{\theta\text{old}}}\ (\pi_\theta) \leq \delta
\end{equation}
where $\theta_{\text{old} }$ is the vector of policy parameters before the update. The KL-divergence constraint is used to make sure the updated policy does not move too far away from the current policy. TRPO can guarantee a monotonic improvement over policy iteration. We refer to  \cite{DBLP:journals/corr/SchulmanWDRK17} for more details.  

In addition, PPO uses a clipping operation to improve the learning. The new objective function with clip is defined as follows:  
\begin{equation}
    L^{CLIP}(\theta) =  \mathbb{\hat{E}}_t \Big[\min(r_t(\theta) \hat{A}_t,\text{clip}(r_t(\theta), 1-\epsilon, 1+\epsilon) \hat{A}_t ) \Big]
\label{Equation:objectivePPO}
\end{equation}  
where $ r_t (\theta) =  \frac{\pi_\theta (a_t|s_t)}{\pi_{\theta\text{old}} (a_t|s_t)}$ is a ratio between the new policy and the old policy that  
measures how different the two policies are. In this way, the objective function clips the estimated advantage function if the new policy is far away from the old one. 
The algorithm computes the expectation over batches of samples and the expectation operator is taken over the minimum of two terms. 
The first of these terms in  $ r_t(\theta)$ multiplied with the advantages estimate  $\hat{A}_t$. This is the default objective for normal policy gradients which pushes the policy towards actions that yield a high positive advantage over the baseline. The second term is very similar to the first term by applying a clipping operation between $1-\epsilon$ and $1+\epsilon$, where $\epsilon$ is typically set to somewhere between 0 and 0.2. Then finally, the min operator is applied to the two terms to get the final expectation. 

We use a deep neural network architecture with shared parameters for both policy and value functions. In this case, a combined loss function is applied: 
 \begin{equation}
     L_t^{CLIP+VF+S}(\theta) = \mathbb{\hat{E}}_t \Big[L_t^{CLIP}(\theta) - c_1L_1^{VF}(\theta)\ +\ c_2S[\pi_\theta](s_t) \Big]
 \label{Equation:lossfunctionppo}
 \end{equation} 
This final loss function is used to train the agent. In addition to the clipped PPO objective function (\ref{Equation:objectivePPO}), the loss function has the additional loss function $L_1^{VF}(\theta)$, which is a squared-error loss $(V_\theta(s_t)-V_{\text{target}})^2$. It is in charge of updating the baseline network, 
determining what the discount reward will be over the run while being in state $s_t$. The second additional term $S$ denotes an entropy bonus. It encourages exploration. $c_1$ and $c_2$ are hyperparameter constants to weigh the contributions of these terms. 

To successfully train the PPO algorithm that learns which action to take in a given state, a training algorithm is created, see 
Algorithm \ref{Algorithm:PPO}. The algorithm starts by initializing the environment by creating a connection with the simulation model. Then the first state is received with orders and initial capacities. Given this state, the agent estimates the advantage $\hat{A}_t$. Each step the agent performs an action that is provided by the policy. Feasible actions are sent to the simulation model and simulated in the environment. While the action is not feasible, the agent keeps on predicting until a feasible action is chosen. Subsequently, the action is simulated in the environment and possible reward $r_t$ is received by the agent in the meantime. After the state has changed, the simulation model provides the next state $s_{t+1}$. Based on $s_t, s_{t+1}, r_t$, the policy is updated by maximizing the 
PPO objective, via stochastic gradient ascent with Adam \cite{kingma2014adam}.

\begin{algorithm}[H]
 PPO Initialization\\
 Simulation model initialization create interaction between agent and model\\
 \For{episode = 1 to T-steps}{
  Load dataset set and initialize problem instance\\
  sample action $a_t$ using current policy $\pi_\theta$\\
  \eIf{Action == Feasible}{

    Let environment simulate action $a_t$ and receive reward $r_t$ and the next state $s_{t+1}$\;
   }{
    Receive penalty $r_t$ for infeasible action\;
    $s_{t+1} = s_{t}$\;
  }
Compute advantage estimates $\hat{A}_t$,...., $\hat{A}_T$\\
Optimize $L_t$ wrt $\theta$, via minibatch gradient descent\\
$\pi_\theta$ = $\theta_{\text{old} }$ \\
\If{end of episode}{
 break\;
}
  }
 \caption{The PPO algorithm for OBSP}\label{Algorithm:PPO}
\end{algorithm}

\section{Experimental evaluation}\label{Section: Experiment Setup}
This section describes the experimental evalution of the proposed DRL approach for solving the OBSP. First, we present the experimental setup and define two scenarios based on problem instances from practice. Second, we propose several heuristic rules for the OBSP, which are used as performance comparisons to our DRL approach. Finally, we present and discuss the experiment results.

\subsection{Experimental setup}
The experiment instances are derived based on a dataset from practice. We used a data set that contains 257.585 orders with a total of 376.522 items. Moreover, we took samples from the dataset to vary the parameters of which we expect that they have an impact on the performance of the algorithm. In particular, we vary the following parameters.
\begin{enumerate}
    \item \textbf{Order throughput.} Throughput rates between 300 orders per hour and 500 orders per hour are analyzed. 
    \item \textbf{Resource settings.} In the resource settings the number of pickers, shuttles, StO workstations, DtO workstations, and Packing workstations can be adjusted. These settings are adjusted to the desired throughput rate such that the tardy orders are minimized.
    \item \textbf{Distribution of SKUs in storage systems.} In the setup of the experiment, the percentage of orders that are picked within the PtG area is kept at 70\% and 30\% in the GtP area. In the data that is used for this experiment, 80\% of the requested orders can be delivered by approximately 20\% of the SKUs, these are the fast movers. The fast movers are therefore stored in the PtG area and the slow movers are stored in GtP. Then during peak days, additional operators can be easy deployed in the PtG area to pick the fast movers. Whereas deploying additional shuttles in the GtP area is not that flexible. Therefore, slow movers are stored in the GtP area and thus 30\% of the orders are picked in this area; 20\% slow movers and 10\% fast movers. These ratios can be adjusted to satisfying needs however this is left out of scope in this paper.
    \item \textbf{Length of run.} The length of a run that is analyzed is set to a simulation run time of 60 minutes. In these 60 minutes, we analyze what the number of tardy orders is for a certain throughput with specified resource settings. Consequently, only the hours that include cut-off times are analyzed in the settings of the experiment, i.e. every hour before 19:00 is not included. 
    \item \textbf{Order releasing moments.} We examine two settings for releasing orders into the system. Orders can be released every 60 minutes or every 15 minutes. When including more release moments per hour, a more dynamic environment is tested.
    \item \textbf{Cut-off moments.} We also examine two settings with different cut-off times, similar to the order releasing setting.   cut-off times can occur every; 60 minutes or every 15 minutes. In the first setting only five cut-off times occur, i.e. 20:00, 21:00, 22:00, 23:00 and 24:00. In the second setting, every 15 minutes orders are shipped, resulting in 17 cut-off times between 20:00 and 24:00.
    \item \textbf{Items per order.} For our dataset the distribution of the number of items per order is shown in Table~\ref{table:Distribution of items}. We use this distribution for our experiments. 
\end{enumerate}

\begin{table}[h]
\centering
\footnotesize{
\begin{tabular}{|l||l|l|l|l|l|l|l|l|l|l|}
\hline

\textbf{Nr. of items} & \textbf{1} & \textbf{2} & \textbf{3} & \textbf{4} & \textbf{5} & \textbf{6} & \textbf{7} & \textbf{8} & \textbf{9} & \textbf{10} \\ \hline

Orders (\%)                    & 74\%    & 16\%    & 5\%     & 2\%     & 1\%     & 1\%     & 0.2\%     & 0.2\%     & 0.2\%     & 0.2\%      \\ \hline
\end{tabular}
}
\caption{Distribution of items per order}
\label{table:Distribution of items}
\end{table}

We derive two scenarios based on these parameters: a high dynamicity planning scenario and a low dynamicity planning scenario, i.e.
\begin{enumerate}
\item[] \textbf{Scenario A (low Dynamicity)}.  
In this scenario, we set the length of run to 1 hour, the cut-off moments to every 60 minutes, the order releasing moment to every 60 minutes, and the distribution of SKUs to 70\% in PtG and 30\% in GtP.  
\item[] \textbf{Scenario B (low Dynamicity)}.
We set the length of run to 1 hour, the cut-off moments to every 15 minutes, the order releasing moment to every 15 minutes and the distribution of SKUs to 70\% in PtG and 30\% in GtP.
\end{enumerate}
Using experiments, scenarios A and B are analyzed for different resource settings and different order throughput times, during several hours between 19:00 and 23:00. The order throughput determines the size of the instance solved. 
Tables  \ref{table:instance} and \ref{Table:Dynamicinstance} show examples of the generated problem instances for Scenario A (with 300 orders) and Scenario B (with 330 orders), respectively.

\begin{table}[ht]
\centering
\begin{tabular}{l|l|l|l|l|l|l|}
\cline{2-7}
{\textbf{}}                                    & \multicolumn{2}{c|}{\textbf{PtG}}                                     & \multicolumn{2}{c|}{\textbf{GtP}}                                     & \multicolumn{1}{c|}{\textbf{Both}} & \multicolumn{1}{c|}{} \\ 
\cline{2-6}  & \multicolumn{1}{c|}{\textbf{SIO}} & \multicolumn{1}{c|}{\textbf{MIO}} & \multicolumn{1}{c|}{\textbf{SIO}} & \multicolumn{1}{c|}{\textbf{MIO}} & \multicolumn{1}{c|}{\textbf{MIO}} &
\multicolumn{1}{c|}{\textbf{Total}}
\\ \hline
\multicolumn{1}{|l|}{\textbf{N orders}}                         & {149}        & {46}         & {72}         & 17                                & 16                                     & 300                                                   \\ \hline
\multicolumn{1}{|l|}{\textbf{N items}}                          & 149                               & 131                               & 72                                & 55                                & 38                                     & 445                                                   \\ \hline
\multicolumn{1}{|l|}{\textbf{COT* $\leq$} \textbf{60}}    & 30.00\%                           & 12.00\%                           & 14.00\%                           & 3.33\%                            & 0\%                                    & 59.33\%                                               \\ \hline
\multicolumn{1}{|l|}{\textbf{COT* \textgreater 60}} & 19.67\%                           & 3.33\%                            & 10.00\%                           & 2.33\%                            & 5.33\%                                 & 40.67\%                                               \\ \hline
\end{tabular}
 \caption{Summary of problem instance example for scenario A with a throughput of 300 orders. COT denotes cut-off times (in minutes).}
  \label{table:instance}
\end{table}

\begin{table}[h]
\centering
\begin{tabular}{l|l|l|l|l|l|l|}
\cline{2-7}
{}                                          & \multicolumn{2}{c|}{\textbf{PtG}}                                     & \multicolumn{2}{c|}{\textbf{GtP}}                                     & \multicolumn{1}{c|}{\textbf{Both}} & \multicolumn{1}{c|}{}                                 \\ \cline{2-6}  & \multicolumn{1}{c|}{\textbf{SIO}} & \multicolumn{1}{c|}{\textbf{MIO}} & \multicolumn{1}{c|}{\textbf{SIO}} & \multicolumn{1}{c|}{\textbf{MIO}} & \multicolumn{1}{c|}{\textbf{MIO}}      & 
\multicolumn{1}{c|}{\textbf{Total}} \\ \hline
\multicolumn{1}{|l|}{\textbf{N orders}}                               & {161}        & {59}         & {69}         & 29                                & 12 & 330                                                   \\ \hline
\multicolumn{1}{|l|}{\textbf{N items}}                                & 161                               & 160                               & 72                                & 85                                & 26                                     & 504                                                   \\ \hline
\multicolumn{1}{|l|}{\textbf{COT* $\leq$  \text{15}}  }               & 24.24\%                           & 10.91\%                           & 4.55\%                            & 3.33\%                            & 3.64\%                                 & 47\%                                                  \\ \hline
\multicolumn{1}{|l|}{\textbf{15 \textgreater COT* \text{$\leq$  30}}} & 15.15\%                           & 2.73\%                            & 6.97\%                            & 2.73\%                            & 0.00\%                                 & 28\%                                                  \\ \hline
\multicolumn{1}{|l|}{\textbf{30 \textgreater COT* \text{$\leq$ 45}}}  & 6.06\%                            & 3.94\%                            & 5.15\%                            & 0.91\%                            & 0.00\%                                 & 16\%                                                  \\ \hline
\multicolumn{1}{|l|}{\textbf{COT* \textgreater  45}}               & 3.33\%                            & 0.30\%                            & 4.24\%                            & 1.82\%                            & 0.00\%                                 & 10\%                                                  \\ \hline
\end{tabular}

\caption{Summary of problem instance example for scenario B with a throughput of 330 orders. COT denotes cut-off times (in minutes)}
\label{Table:Dynamicinstance}
\end{table}

\subsection{Training Parameters} 
The model parameters for this experiment consist of parameters to train the PPO algorithm and the values of these parameters are similar to the original work of  \cite{DBLP:journals/corr/SchulmanWDRK17}. However, some adjustments have been made to fit the OBSP environment. The algorithm is set to train for $750.000$ steps, depending on the size of the problem instance, the episode requires 50 up to 100 actions.

We initialize a Neural Network similar to \cite{DBLP:journals/corr/SchulmanWDRK17} with two hidden layers of 64 units, and tanh nonlinearities, outputting the mean of a Gaussian distribution for our action space of 31 units. The clipping parameter that showed the best performance in \cite{DBLP:journals/corr/SchulmanWDRK17} was set to 0.2. This ensures that the updated policy cannot differ from the old policy by 0.2. The discount factor $\gamma$ is set to $0.9999$ instead of $0.99$. This means that we care more about reward that is received in the future than immediate reward. Both the DRL agent and the simulation model are utilized on the same processing machine with an Intel(R) Core(TM) i7 Processor CPU @ 2.80GHz and 32GB of RAM.

\subsection{The proposed heuristics for the OBSP} 
For the purpose of comparisons, 
we propose several batching and sequencing heuristics. These heuristics are designed based on the existing literature (see Section\ref{Subsection:Literature OBSP}) and discussions with several warehouse setup designers. The proposed heuristics are described and implemented in a 2-step approach. In the first step orders are batched and in the second step the batches or orders are sequenced. The following 4 batching rules are examined.

\textbf{The greedy batching algorithm}  
creates an initial solution by sorting the orders in ascending order of cut of times and selecting the first order as first solution. In case of an GtP order, no orders are added to the initial solution. GtP orders are always picked-by-order so the initial solution is also the complete solution. PtG orders are processed in a batch and orders are added to the solution as long as the maximum batch size is not violated. Maximum batch size for a SIO and MIO batch is 10 and 12 respectively.

\textbf{The least slack time batching rule (LST)} is an extension of the greedy batching algorithm. Batches in PtG are formed by the greedy algorithm. However, the Least Slack Time batching rule can dismantle the batch into separate orders once the slack time of the batch becomes less than zero. Picking-by-order is then performed. Slack time is the amount of time left until its due date of the order is reached when also taking into consideration the processing time that is required to pick and consolidate the order/batch. 

\textbf{The picking-by-order small batches (POSB)} is  also  an  extension  rule  to  the  greedy  algorithm  and is created for batches with four or fewer orders. When a batch consist of 4 or less orders in the PtG area, picking-by-order is applied instead of picking-by-batch. This is possible since a picker can carry a maximum of four carton boxes on a picking cart. 

Finally, \textbf{combining LST and POSB batching rules} is a combination of the LST and POSB batching rule. 

After the orders and batches have been created, the orders can be sequenced with the following 5 sequencing rules.

\textbf{The Earliest Due Date (EDD) rule} sequences all orders based on cut-off time i.e. earliest due dates. The greedy algorithm already does this for releasing orders and batches to both storage systems. However, batches that are transported from PtG to GtP, are placed at the end of the queue.  This is the same queue in which all the orders wait that only require GtP picking. In case that these batches have an earlier due date than some GtP orders, the batches with the earlier due date are then again sequenced such that these are released and picked earlier than the batches/orders that are awaiting in the GtP area.

\textbf{The Longest Processing Time (LPT) rule} sequences orders based on processing times. The processing time in this case is the total picking time required for the order or batch. 

\textbf{The Shortest Processing Time sequencing (SPT) rule } is the opposite of the LPT rule and therefore leaving all the batches with the longest picking time for the very last end.

\textbf{The Maximum Total Processing Time sequencing (MAXTP) rule} sequences the orders in descending order of the MAXTP. In this rule the consolidation time at the workstations is also included. 

\textbf{The Least Slack Time sequencing (LST) rule} assigns priority based on the slack time of a process. A large proportion of the orders have similar cut-off time, however, when the LPT or the MAXTP is applied, more priority could be assigned to orders that do have a later due date time. The LST does not provide this priority to these orders but provides this priority to the orders that have a longer processing time but also have to be shipped earlier. Orders with the Least Slack Time are processed first.

\subsection{The experimental results}
The objective of OBSP is to minimize the number of tardy orders. In the result section, we report the performance of the DRL approach and proposed heuristics in terms of the number of tardy orders that is produced for a particular experiment setup. 

First, we test the performance of the proposed batching and sequencing heuristics, which is summarized in Table \ref{Table:Batching_and_sequencing_rules}. This table shows that the LST batching rule and the LST+POSB batching rule outperform the other batching rules. In terms of the sequencing rules, the LST is  superior to the other sequencing rules. Consequently, for benchmarking purposes, the LST+POSB batching rules in combination with the LST sequencing rule is used to compare the performance of the PPO algorithm. 

\begin{table}[h]
\centering
\begin{tabular}{|l|l|c|c|c|c|}
\hline
 & Setting & GR & LST & POSB & LST+POSB \\ \hline
 \multirow{2}{*}{EDD} & 300-3-10-1-1-1 & 9.33\%  & 5.67\%  &  8.57\%  &  5.17\% \\ 
 \cline{2-6} 
    &  300-5-10-1-1-1  &   1.41\% &  0.72\% &  0.76\% &  0.75\%  \\ \hline
 \multirow{2}{*}{LPT} & 300-3-10-1-1-1 & 8.04\% &  6.90\%  &  7.38\%  &  7.53\%  \\
 \cline{2-6} 
    &  300-5-10-1-1-1  & 1.36\% & 1.58\% &  1.83\%  & 2.32\% \\ \hline
 \multirow{2}{*}{SPT} & 300-3-10-1-1-1 &   24.78\% &  5.95\% &  22.86\% & 7.15\% \\
 \cline{2-6} 
    &  300-5-10-1-1-1  &   14.37\% & 12.46\% & 15.17\% & 11.01\% \\ \hline
 \multirow{2}{*}{MAXTP} & 300-3-10-1-1-1 & 5.87\% &  6.24\% &  7.29\% &  5.98\% \\
 \cline{2-6} 
    &  300-5-10-1-1-1  &  0.80\% & 1.08\% & 1.37\% & 1.56\%  \\ \hline
 \multirow{2}{*}{LST} & 300-3-10-1-1-1 & 7.91\% & 5.88\% &  9.03\% &  4.76\% \\
 \cline{2-6} 
    &  300-5-10-1-1-1  &    0.24\%  & 0.06\% & 0.07\%  &  0.04\% \\ \hline
\end{tabular}
\caption{The average percentage of tardy orders for four batching rules (each column indicating a batching rule) in combination with five sequencing rules (each row indicating a sequencing rule). In each run a throughput between 19:00 and 20:00 for scenario A is evaluated and the results are averaged over 3000 runs. Setting: either 300 orders and 3 pickers (300-3-10-1-1-1) or 300 orders and 5 pickers (300-5-10-1-1-1). Both settings have 10 shuttles and 1 workstation of each type.}
\label{Table:Batching_and_sequencing_rules}
\end{table}

Table \ref{table:overview_PPO_LST} shows the performance of our PPO algorithm (expressed as the percentage of tardy orders) compared to the performance of the LST batching rule in combination with the LST+POSB sequencing rule.

The PPO algorithm is trained on the training data which accounts for 70\% of the original data. Subsequently, to test performance, the agent is tested on the test set. The used training parameters are similar to the work of  \cite{DBLP:journals/corr/SchulmanWDRK17}, but modifications have been made to fit the OBSP environment. The algorithm is set to train for $750.000$ steps. The number of steps to run per update is set to 1025 steps and the discount factor $\gamma$ is set to $0.9999$.

Table \ref{table:overview_PPO_LST} shows that the PPO outperforms the heuristics for scenario A, but that the heuristics show superior results for smaller instances compared to the PPO for scenario B. The PPO outperforms the heuristics in performance for the 500 instance size in scenario B. However, these results are not significant, especially with the a standard deviation of 2.89\%. The results are obtained by training four agents that show similar training behavior. After training, one of the agent is used for testing to obtain the average results over 100 episodes. 

\begin{table}[h]
\centering
\begin{tabular}{l|c|c|c|c|}
\cline{2-5}
                                             & \multicolumn{2}{c|}{\textbf{Scenario A}} & \multicolumn{2}{c|}{\textbf{Scenario B}} \\ \hline
\multicolumn{1}{|l|}{\textbf{Instance size}} & \textbf{LST}      & \textbf{PPO}         & \textbf{LST}      & \textbf{PPO}         \\ \hline
\multicolumn{1}{|l|}{\textbf{330-5-8-1-1-1}} & 0.96\%            & 0.54\% (1.59\%)      & 0.22\%            & 2.68\% (1.56\%)      \\ \hline
\multicolumn{1}{|l|}{\textbf{400-5-8-1-1-1}} & 4.07\%            & 1.08 (2.00\%)        & 1.57\%            & 3.32\% (2.06\%)       \\ \hline
\multicolumn{1}{|l|}{\textbf{500-5-8-1-1-1}} & 11.67\%           & 3.66 (2.59\%)        & 4.52\%            & 4.18\% (2.89\%)      \\ \hline
\end{tabular}

\caption{The average percentage of tardy orders for the DRL and LST heuristic (LST batching and combination of LST and POSB for sequencing) for Scenario A and B. The results of the DRL are the mean and standard deviation (between brackets) over a training duration of 1000 episodes. Three settings; 330, 400 and 500 orders with each 5 pickers, 8 shuttles and 1 workstation of each type.}
\label{table:overview_PPO_LST}
\end{table}

Why is the PPO superior in scenario A and not superior in scenario B? On basis of the strategy analysis, we found that the PPO applies a different strategy for scenario A compared to the heuristics. Whereas the heuristics apply picking-by-order is the GtP area, the PPO prefers applies both picking strategies: picking-by-order and picking-by-batch. It can be concluded that by the use of DRL, new strategy insights have been provided which show to be beneficial to the OBSP. Initially, this strategy was found to be disadvantageous and was therefore not applied in the heuristics. However, the DRL approach showed the opposite. 

When comparing the strategy of agents of scenario A with the strategy of agents of scenario B, we can conclude that the agents for scenario B apply a more varied strategy. Agents in scenario B performs more picking-by-order actions than the agents for scenario A. Agents for scenario B in total include 22 actions of the 31 actions into their strategy. Whereas the agents for scenario A eventually only consider 9 actions in their strategy. This is a result of more cut of times for scenario A. In this scenario there are more critical orders than in scenario B and therefore more individual orders are picked-by-order. As a result, more actions are considered by the agents for scenario B.    

\begin{figure}[h]
  \includegraphics[width=\linewidth]{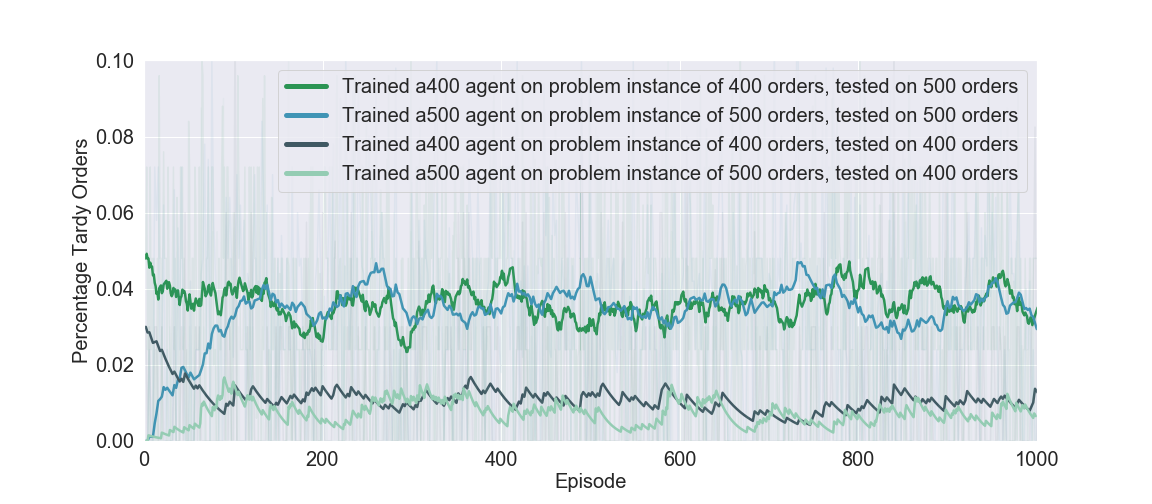}
  \caption{Percentage of tardy orders for different trained agents on Scenario A.}
  \label{fig:TardyOrders_difagents}
\end{figure}

Furthermore, we investigated the generalizability of the DRL approach with two different cases; two trained agents for specific instance size problems are tested on different instance sizes and a trained agent on a fixed time period (i.e. from 19:00 to 20:00) is tested on a random time period (somewhere between 19:00 and 00:00). For the first cases, we trained two agents for scenario A; $a400$ and $a500$. These two agents considered a problem instance of 400 and 500 orders respectively during training. For this case, the trained $a400$ agent is tested on a problem instance with a throughput of 500 orders, and vice versa; the trained $a500$ agent is tested on a problem instance with a throughput of 400 orders. Results in terms of the percentage of tardy orders are shown in Figure \ref{fig:TardyOrders_difagents}.

It can be concluded that both agents that are tested on unfamiliar instances, perform as equally as the agents that are trained on these instances. The trained agents show approximately the same percentages of tardy orders and standard deviations, see Table \ref{table:diftrainedagents} for a complete overview.

\begin{table}[h]
\centering
\begin{tabular}{|l||l|l|}
\hline
\textbf{Instance size} & \multicolumn{1}{c|}{\textbf{a400}}     & \multicolumn{1}{c|}{\textbf{a500}} \\ \hline
\textbf{400 orders}          & {1.08\% (2.00\%)} & {0.75\% (1.68\%)}            \\ \hline
\textbf{500 orders}          & 3.57\% (2.52\%)                        & 3.66\% (2.59\%)                    \\ \hline
\end{tabular}
\caption{The average percentage of Tardy orders for the a400 agent and the a500 agent on an instance size 400 and 500 orders. The results are the mean and standard deviation (between brackets) over a training duration of 1000 episodes}
\label{table:diftrainedagents}
\end{table}

\begin{figure}[ht]
  \includegraphics[width=\linewidth]{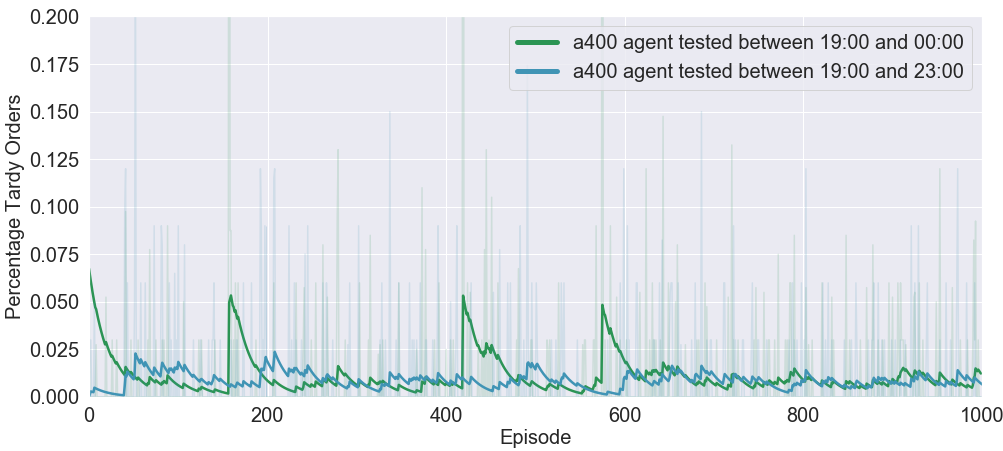}
  \caption{Performance of two trained agent (a400) on problem instance of 400 orders, tested on random hours of 400 orders}
  \label{fig:TardyOrders_randomhours}
\end{figure}

For the second case, we train an agent on an problem instance that starts at 19:00 and then test this agent on random hours between 19:00 and 00:00. Figure \ref{fig:TardyOrders_randomhours} shows the performance of the $a400$ agent that is tested on random hours between 19:00 and 00:00 and random hours between 19:00 and 23:00. It can be seen that the $a400$ does not indicate a stabilized performance during all episodes when the last hour is also included. After approximately 180, 410 and 590 episodes the agent does not seem to be able to solve the instance and many orders become tardy. However, when excluding the last hour, generalization of the DRL approach is improved. In the instances that include the last hour, we process the last remaining orders that have to be shipped just before 24:00. However, it can be the case that only 100 to 200 orders are requested during this hour. As a result of a significantly smaller instance, the agent could end up in states that the agent has not encountered during training. The overall performance is nevertheless similar, with an average percentage of tardy orders of 0.66\% and 0.94\% for the agent that included and excluded the last hour respectively.

\section{Conclusion}\label{Section:Conclusions}
In this work, we have shown that benefits from recent advancements in Deep Reinforcement Learning (DRL) in solving a challenging  decision-making problem in warehousing, i.e. the order batching and sequencing problem. 
We apply the Proximal Policy Optimization (PPO) algorithm 
by first transferring the OBSP to a Semi-Markov Decision Process formulation. Based on this formulation, we create a DRL agent that takes OBSP decisions and interacts with a simulated environment that calculates the effects of these decisions. We have compared the performance of the DRL approach to that of several proposed heuristics with a set of experiments with 
with a variety of settings. The experimental results have demonstrated that the DRL agent is able to find good strategies to solve the OBSP for all different tested settings and outperforms the heuristics for most of the tested settings. In addition, the experiment results demonstrate some level of robustness and generalizability of the DRL approach. More specifically, the agents 
are trained on a particular set of orders (e.g. 400 orders) but the learned strategies can produce good results on different settings with less, or more orders. 
Moreover, the agents trained to handle the orders between 19:00 and 20:00 are able achieve consistent good performance when being applied to different hours.



We have shown in this paper a successful application of the deep reinforcement learning in solving the OBSP. 
Deep reinforcement learning remains a promising approach to solve sequential decision making problems with complex, dynamic environments and is currently well suited to solve problems where no existing strategy is satisfactory, not only in the context of the OBSP, but also many real-world decision making problems that are extensively studied in the Operations Research and Management community. 



\end{document}